\documentclass[letterpaper, 10 pt, conference]{ieeeconf}  
\usepackage{times}
\usepackage{epsfig}
\usepackage{graphicx}
\usepackage{amsmath}
\usepackage{amssymb}
\usepackage{multirow}
\usepackage{microtype}
\usepackage{graphicx}
\usepackage{caption}
\usepackage{subcaption}
\usepackage{cleveref}
\usepackage{soul}
\usepackage[colorlinks]{hyperref}

\IEEEoverridecommandlockouts                              

\title{\LARGE \bf
Conf-Net: Toward High-Confidence Dense 3D Point-Cloud with Error-Map Prediction
}
\author{Hamid Hekmatian\\
{\tt\small hamid.hekmatian@havalus.com}
\and
Jingfu Jin\\
{\tt\small jingfu.jin@havalus.com}
\and
Samir Al-Stouhi\\
{\tt\small samir.alstouhi@havalus.com}
}

\begin{document}

\maketitle
\thispagestyle{empty}
\pagestyle{empty}

\begin{abstract}
      This work proposes a method for depth completion of sparse LiDAR data using a convolutional neural network which can be used to generate semi-dense depth maps and "almost" full 3D point-clouds with significantly lower root mean squared error (RMSE) over state-of-the-art methods. We add an "Error Prediction" unit to our network and present a novel and simple end-to-end method that learns to predict an error-map of depth regression task. An "almost" dense high-confidence/low-variance point-cloud is more valuable for safety-critical applications specifically real-world autonomous driving than a full point-cloud with high error rate and high error variance. Using our predicted error-map, we demonstrate that by up-filling a LiDAR point cloud from 18,000 points to 285,000 points, versus 300,000 points for full depth, we can reduce the RMSE error from 1004 to 399. This error is approximately \textbf{60\% less} than the state-of-the-art and 50\% less than the state-of-the-art with RGB guidance (we did not use RGB guidance in our algorithm).
      In addition to analyzing our results on Kitti depth completion dataset, we also demonstrate the ability of our proposed method to extend to new tasks by deploying our "Error Prediction" unit to improve upon the state-of-the-art for monocular depth estimation. Codes and demo videos are available at \url{http://github.com/hekmak/Conf-net}.
\end{abstract}

\begin{figure*}
    \centering
    \begin{subfigure}[b]{0.35\textwidth}
        \centering
        \includegraphics[width=0.9\textwidth]{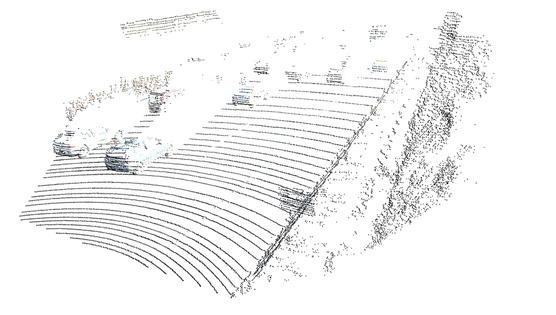}
        \caption{Raw LiDAR}
        \label{results_1:a}
    \end{subfigure}%
    \begin{subfigure}[b]{0.35\textwidth}
        \centering
        \includegraphics[width=0.9\textwidth]{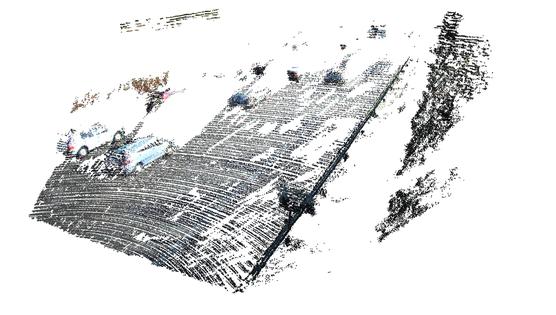}
        \caption{Ground truth}
        \label{results_1:b}
    \end{subfigure}%
    \begin{subfigure}[b]{0.35\textwidth}
        \centering
        \includegraphics[width=0.9\textwidth]{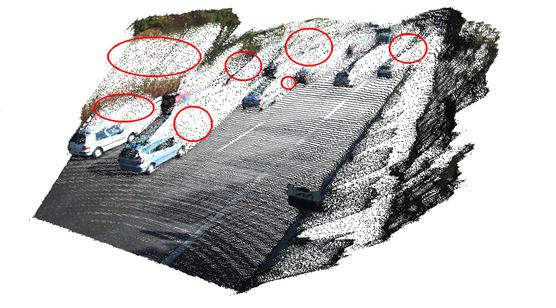}
        \caption{Predicted point-cloud}
        \label{results_1:c}
    \end{subfigure}%
    \hfill 
    \begin{subfigure}[b]{0.35\textwidth}
        \centering
        \includegraphics[width=0.9\textwidth]{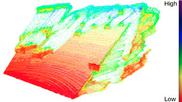}
        \caption{\begin{tabular}{@{}c@{}}Predicted point-cloud\\ colored with our predicted error-map\end{tabular} }
        \label{results_1:d}
    \end{subfigure}%
    \begin{subfigure}[b]{0.35\textwidth}
        \centering
        \includegraphics[width=0.9\textwidth]{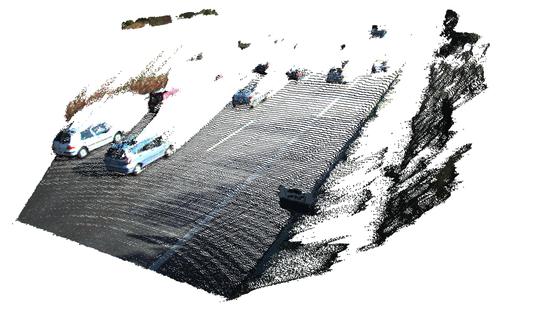}
        \caption{\begin{tabular}{@{}c@{}}High-confidence\\ point-cloud\end{tabular}}
        \label{results_1:d}
    \end{subfigure}%
    \caption{Illustration of depth completion process on 3D point-clouds (RGB values were used only for coloring). Using a raw 3D point-cloud (a), we predict a dense point-cloud (c) along with the error-map (d), which we use to acquire our final high-confidence point-cloud (e).}
    \label{results_1}
\end{figure*}
\section{INTRODUCTION}
Cost-effective LiDAR sensor with high resolution and dense point-cloud will play a crucial role in commercializing autonomous vehicles. So far, most of the commercialized LiDAR sensors generate sparse point-cloud due to the limited number of beams. The cost of the sensor increases substantially with the increased number of beams and resolution. Depth completion techniques are an inexpensive algorithmic aid to fill the sparsity of LiDAR point clouds. Various applications (e.g., object detection, visual odometry, SLAM, etc.) can benefit from a more dense point-cloud. 

Sparse depth-map is a projection of 3D LiDAR point-cloud into 2D image space where the value of each pixel in 2D image space corresponds to the depth of each point. We usually define dense depth as a depth-map which fully covers the RGB camera space. Datasets like Kitti depth completion \cite{uhrig2017sparsity} generate their ground-truth depths by enforcing consistency between accumulated LiDAR scans and stereo depth (generated by stereo cameras), so even without RGB guidance depth completion is done in 2D camera space. Densifying sparse depth maps lies in the field of depth completion which also has been studied as depth super-resolution \cite{song2016deep}, depth enhancement \cite{zhang2016fast,lu2014depth} and depth in-painting \cite{miao2012texture}.

In this paper, we focus on LiDAR-based depth completion without RGB guidance. We did not use RGB guidance in our algorithm since it is not available for 360$^\circ$, only covers the front view and additionally is not reliable during night time or bad weather. We have designed a novel method which makes networks able to predict their own error-map, we call the network Conf-Net. We achieve this by using the pixel-wise error in each training step as an extra learning signal for training our neural network. We directly learn both dense depth and its corresponding error-map in a multi-tasking manner. 2D dense depth along with the predicted error-map are used to generate a high-confidence 3D dense point-cloud.

In LiDAR-based depth completion, error and uncertainty increase substantially due to high sparsity in input data (more than 95\% sparsity in the input image). The sparsity makes it not possible to accurately predict the depth on far away areas, areas with high sparsity ad areas with abrupt change in depth values (e.g., edges), especially without RGB guidance. To the best of our knowledge, this is the first paper predicting the pixel-wise error-map of a convolutional neural network for depth completion task. The approaches \cite{eldesokey2018confidence,kendall2018multi} predict the confidence and the uncertainty maps but all of these methods deploy an indirect way of estimating confidence (or uncertainty), in contrast our approach learns to estimate the error-map directly from the data. 

To demonstrate that our proposed method is applicable to other regression tasks, we integrate our "Error Prediction" unit in a monocular depth estimation network on NYU Depth dataset \cite{silberman2012indoor}. We demonstrate that our approach can be easily added to existing networks by only modifying the loss function while other existing confidence based approaches \cite{eldesokey2018propagating,eldesokey2018confidence,xia2019generating} require re-implementing the whole network. Our main contributions are listed below:
\begin{enumerate}
\item We propose a novel method to predict a high-quality pixel-wise error-map. Our approach outperforms similar existing methods in terms of uncertainty and confidence. 
\item We used our error-map to generate high confidence 3D full point-cloud with low error variance from sparse LiDAR point-cloud. Our point-cloud is 15 times more dense than input (which is Velodyne HDL-64) and 3 times more accurate than the state-of-the-art (RMSE = 300mm). The process is shown in Figure~\ref{results_1}.
\item We conduct an uncertainty based analysis of Kitti depth completion dataset.
\item We demonstrate that our method can be easily adapted to other regression tasks and existing networks by predicting the error-map of monocular depth estimation on NYU Depth dataset. 
\end{enumerate}

\section{Related Works}
\subsection{Depth Completion}
Authors in \cite{uhrig2017sparsity} provide a baseline on Kitti depth completion dataset using Nadaraya-Watson kernel regression. Aditionally, they formulate depth completion as a deep regression problem and solve it end-to-end. They propose sparsity invariant convolution, which uses the validity mask of the sparse depth for a better result. Authors in \cite{huang2018hms} extend the idea of sparsity invariant convolution proposed in \cite{uhrig2017sparsity} by designing a sparse invariant up/down sampling blocks. Using compressed sensing and alternating direction neural networks (ADNNs), authors in \cite{chodosh2018deep} predict the dense depth with fewer parameters compared to other deep models. In \cite{ma2018self}, authors deploy an encoder-decoder architecture with skip layers and residual blocks to reconstruct the depth while \cite{jaritz2018sparse} approaches the depth-completion using an encoder-decoder scheme based on NASNet \cite{zoph2018learning}. The authors in \cite{chan2008noise, ku2018defense} tackle the depth completion using classical computer vision algorithms running on CPU. 
\vspace{-1.0em}
\subsection{Uncertainty and Confidence}
Uncertainty and confidence in depth completion has been approached in two different categories. Methods in the first category use uncertainty and confidence as an internal process to predict finer dense depth \cite{qiu2018deeplidar,van2019sparse}. Methods in the second category predict a pixel-wise uncertainty or confidence map along with their predicted depth as output \cite{eldesokey2018propagating,eldesokey2018confidence,kendall2017uncertainties}. Authors in \cite{qiu2018deeplidar} and \cite{van2019sparse} use uncertainty as an internal step in their model to achieve better depth completion result. In \cite{qiu2018deeplidar}, using surface Normals and the predicted confidence of the warped depth (generated by LiDAR/camera calibration) as guidance, they predict finer dense depth. In \cite{van2019sparse}, the authors use uncertainty to merge the predicted depth of their RGB and LiDAR networks, which result in superior performance in depth completion. 

To our knowledge, the only methods which predict confidence for their dense depth in LiDAR-based depth completion as an independent output are \cite{eldesokey2018propagating,eldesokey2018confidence}. They predict the confidence using normalized convolution layers and propagate it through the network. They use a loss function which simultaneously maximizes the confidence and minimizes the depth prediction loss. Authors in \cite{kendall2017uncertainties,kendall2018multi} exclusively focus on uncertainty in the context of Bayesian Deep Learning by probabilistic interpretation of deep learning models. They address two different types of uncertainty, Aleatoric uncertainty which captures the noise in observations and Epistemic uncertainty which focuses on model uncertainty (noise inside the model). Although they are not explicitly focusing on depth completion, we find their approach to Aleatoric uncertainty closely related to what we are presenting in this paper. We adopt Aleatoric uncertainty as an alternative approach to our proposed method for predicting high confidence depth completion results.
\vspace{-1.5em}
\section{Approach}
To simultaneously learn a dense depth along with the expected error-map of the model, an end-to-end fully convolutional neural network is developed. Next, we filter the predicted dense depth by setting a threshold value on the predicted error-map. The threshold represents the maximum absolute error tolerance. In this way, the pixels in our predicted dense depth with expected high error are removed. The outcome of this process is a high confidence semi-dense depth map, which we used to generate the final 3D point-cloud. The details of the proposed network architecture for depth completion are explained in this section.

\subsection{Network Architecture of Depth Completion}

\begin{figure}[ht]
    \vspace{-0.5em}
    \begin{center}
        \centerline{\includegraphics[width=\columnwidth]{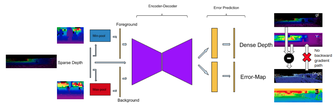}}
        \caption{Our network architecture to learn dense depth and depth's error-map.}
        \label{model_arch}
    \end{center}
    \vspace{-1.5em}
\end{figure}

An overview of the architecture is shown in Figure~\ref{model_arch}. We first estimate the foreground and background pixels which we find it helpful on final dense depth accuracy. Both the foreground and the background maps are concatenated with the sparse input depth. The concatenated maps are then fed into the second step, which is a convolutional neural network with "encode-decoder" architecture. The Encoder-decoder extracts the deep features which we use to predict dense depth and model error simultaneously.
\vspace{-1.0em}

\subsubsection{Pre-processing}
When 3D LiDAR point cloud is projected onto the camera 2D space, the foreground and the background depths get warped due to the displacement between RGB camera and LiDAR sensor. The Kitti depth completion dataset's ground-truth is created by enforcing consistency between the accumulated LiDAR based depth and the depth which results from a stereo reconstruction and removing points with high relative error \cite{uhrig2017sparsity}. Therefore, it only covers the foreground objects. To guide the network to learn areas where the warped transformation occurs we concatenate the estimated foreground and background with the sparse depth. To estimate the background and the foreground, max-pooling and min-pooling operations are applied along with a 15x15 kernel and stride 1.
\subsubsection{Encoder-Decoder}
We use an encoder-decoder architecture with 4 residual blocks (for encoding) and 4 transpose convolution (for decoding) with skip connections, where each block performs strided down-sample/up-sample operation on its input by a factor of 2. The output of this step (32 feature-maps) is fed into our final step to predict the dense depth and the error-map.

\begin{figure*}
	\centering
	\begin{subfigure}[b]{0.2\textwidth}
		\centering
		\includegraphics[width=3.5cm,height=1.25cm]{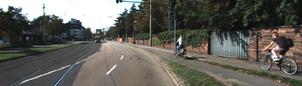}
		\includegraphics[width=3.5cm,height=1.25cm]{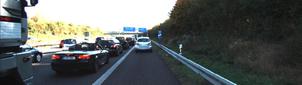}
		
		\caption{RGB}
		\label{fig:gull}
	\end{subfigure}%
	\begin{subfigure}[b]{0.2\textwidth}
		\centering
		\includegraphics[width=3.5cm,height=1.25cm]{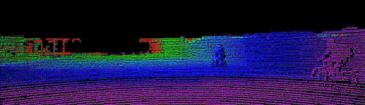}
		\includegraphics[width=3.5cm,height=1.25cm]{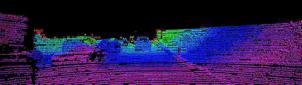}
		
		\caption{LiDAR sparse depth}
		\label{fig:gull}
	\end{subfigure}%
	\begin{subfigure}[b]{0.2\textwidth}
		\centering
		\includegraphics[width=3.5cm,height=1.25cm]{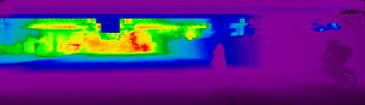}
		\includegraphics[width=3.5cm,height=1.25cm]{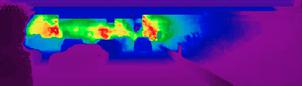}
		
		\caption{Predicted dense depth}
		\label{fig:gull}
	\end{subfigure}%
	\begin{subfigure}[b]{0.2\textwidth}
		\centering
		\includegraphics[width=3.5cm,height=1.25cm]{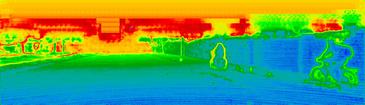}
		\includegraphics[width=3.5cm,height=1.25cm]{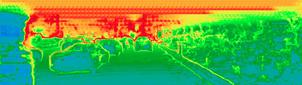}
		
		\caption{Predicted error-map}
		\label{fig:gull}
	\end{subfigure}%
	\begin{subfigure}[b]{0.2\textwidth}
		\centering
		\includegraphics[width=3.5cm,height=1.25cm]{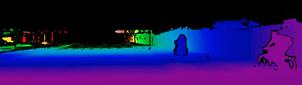}
		\includegraphics[width=3.5cm,height=1.25cm]{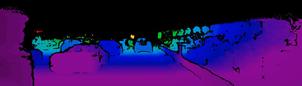}
		
		\caption{High confidence depth}
		\label{fig:gull}
	\end{subfigure}%
	\caption{(a) RGB image (just for visualization purposes), (b) sparse depth (visually enhanced), (c) predicted dense depth, (d) predicted error-map, (e) semi-dense depth after removing points with predicted high errors. Depth values are represented in "nipy spectral" spectrum.}
	\label{var_all}
	
\end{figure*}
\begin{figure*}[ht]
\begin{center}
    \begin{subfigure}[b]{0.2\textwidth}
        \centering
        \includegraphics[width=3.0cm,height=2.5cm]{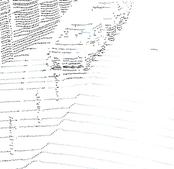}
        \caption{Raw LiDAR}
        \label{results_1:a}
    \end{subfigure}%
    \begin{subfigure}[b]{0.2\textwidth}
        \centering
        \includegraphics[width=3.0cm,height=2.5cm]{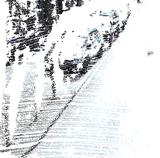}
        \caption{Ground truth}
        \label{results_1:a}
    \end{subfigure}%
    \begin{subfigure}[b]{0.2\textwidth}
        \centering
        \includegraphics[width=3.0cm,height=2.5cm]{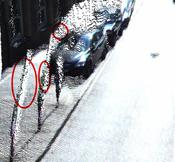}
        \caption{Sparse to Dense}
        \label{results_1:b}
    \end{subfigure}%
    \begin{subfigure}[b]{0.2\textwidth}
        \centering
        \includegraphics[width=3.0cm,height=2.5cm]{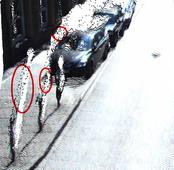}
        \caption{Nconv-L2 (with RGB)}
        \label{results_1:c}
    \end{subfigure}%
    \begin{subfigure}[b]{0.2\textwidth}
        \centering
        \includegraphics[width=3.0cm,height=2.5cm]{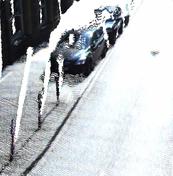}
        \caption{Our final prediction}
        \label{results_1:d}
    \end{subfigure}%

\caption{Close view of the predicted 3D point clouds. Our results look cleaner even from methods who deploy RGB guidance in their prediction. (a) Raw LiDAR point-cloud, (b) ground truth, Sparse to dense \cite{ma2018self}, (c) Nconv-L2 with RGB guidance \cite{eldesokey2018propagating}, (d) our final predicted point-cloud after removing points with predicted high errors.}
\vspace{-1.5em}
\label{close_view}
\end{center}
\end{figure*}
\subsubsection{Error Prediction}
Error prediction is the main unit which makes our approach different from existing methods. In this unit we predict a dense depth along with the error-map of the predicted depth as shown in Figure~\ref{model_arch}. To capture the difference in modalities we add separate streams for depth and error. Depth stream predicts the dense depth and the error stream predicts the depth stream's pixel-wise error. This way, the neural network learns to predict its own pixel-wise error-map. Each prediction has its own loss function, which we denote by $Loss_{depth}$ and $Loss_{error}$. The loss function $Loss_{depth}$ is given by
\vspace{-0.5em}
\begin{equation} \label{RMSE_loss_eq}
    Loss_{depth}=    \frac{1}{w\times h}\sum_{i=1}^{w}\sum_{j=1}^{h} \mid\mid Y_{ij} - \hat Y_{ij} \mid\mid^2
\end{equation}
\vspace{-0.5em}

where $w$ and $h$ represent the width and height of the depth image, and $Y$ and $\hat{Y}$ are the predicted dense depth and depth ground truth, respectively.

For computing the $Loss_{error}$, initially, we need to generate the ground-truth of the error-map. Naively using $|{Y}-\hat{Y}|$ as a ground truth will yield the error stream to predict zero as an error for each pixel. We used a copy operation to overcome this issue.
 To generate the error ground-truth at each training step, we create a copy of $|{Y}-\hat{Y}|$ (which represent pixel-wise absolute error) without backward gradient path and denote it by $gt_{error}$. Afterwards, we treat $gt_{error}$ as a constant and hence we can use it as the ground truth of our error-map to compute $Loss_{error}$. We denote the copy operation by "$\leftarrow$:" as follow
 \vspace{-0.5em}
 \begin{equation} \label{gb_eq}
    gt_{error} \leftarrow:\left |Y - \hat Y  \right | \\
\end{equation}
and define $Loss_{error}$ as the mean squared error between the $gt_{error}$ and $E$ which denotes the network's predicted error, as below 
\begin{equation} \label{gb_loss}
    Loss_{error}:    \frac{1}{w\times h}\sum_{i=1}^{w}\sum_{j=1}^{h} \mid\mid E_{ij} - gt_{error_{ij}} \mid\mid^2 \\
\end{equation}

We update $gt_{error}$ at each training step to represent the new error label. Note that if we do not stop the backward path of the gradient for $gt_{error}$, the network fails to learn the error.  

We also used the idea of copy operation without backward gradient path to normalize our mini-batch losses. At each training step, instead of minimizing the conventional losses, we minimize the ratio of the loss over a baseline. The baseline is a copy of the current loss without backward gradient path. By doing so, we achieved 2 times faster convergence time and slightly better results because of the normalization. Using ratio loss all mini-batches will have same importance no matter what the loss value is, since we minimize the inverse ratio of loss improvement, not its value. We denote the copies of the $Loss_{depth}$ and the $Loss_{error}$ by $baseline_{depth}$ and $baseline_{error}$ as follow
\begin{equation}
    \label{eq_baseline}
    \begin{split}
    baseline_{depth} \leftarrow:  Loss_{depth} \\
    baseline_{error} \leftarrow:  Loss_{error} \\
    \end{split}
\end{equation}
We add normalized values of the losses together as the total loss of our prediction. The formulation is given by
\begin{equation}
    \label{eq_total_loss}
    Loss_{total} = \frac{Loss_{depth}}{baseline_{depth}}+ \frac{Loss_{error}}{baseline_{error}}    
\end{equation}
where $Loss_{total}$ is the final loss function that we send to the optimizer. Ratio loss enables us to predict the error-map without decreasing our model performance regarding RMSE on depth prediction. Losses weighting needed to be manually set as a hyper-parameter without ratio loss. Ratio loss can also be used for other multi-tasking approaches since it enables us to minimize multiple objectives at the same time without the need for tuning the weighting between the loss values. This way we enforce the network to have the same importance for minimizing multiple losses at each training step. We also used this approach on monocular depth estimation in Section IV. 

Figures~\ref{var_all}(a-b) illustrate the RGB image and input sparse depth. Figure~\ref{var_all}(c-e) show the predicted dense depth, predicted error map, and the high confidence semi-dense depth after removing points with predicted error more than 585 mm, respectively. The predicted error-map shown in Figure~\ref{var_all}(d) demonstrates how our network learns the expected depth completion error-map. As we get further from the camera, the network predicts higher error. Predicted error increases significantly on the sky and the areas without depth value in the sparse input. The flat surfaces such as ground show lower errors and edges have large errors. 
\begin{figure*}[ht]
\begin{center}
    \begin{subfigure}[b]{0.3\textwidth}
        \centering
        \includegraphics[width=0.9\columnwidth]{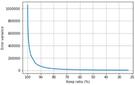}
        \caption{Error variance}
        \label{results_1:a}
    \end{subfigure}%
    \begin{subfigure}[b]{0.3\textwidth}
        \centering
        \includegraphics[width=0.9\columnwidth]{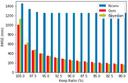}
        \caption{RMSE comparison}
        \label{results_1:a}
    \end{subfigure}%
    \begin{subfigure}[b]{0.3\textwidth}
\centerline{\includegraphics[width=0.7\columnwidth]{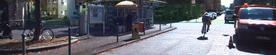}}
\centerline{\includegraphics[width=0.7\columnwidth]{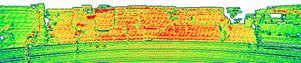}}
\centerline{\includegraphics[width=0.7\columnwidth]{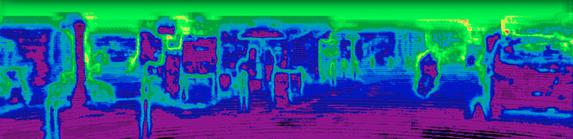}}
\centerline{\includegraphics[width=0.7\columnwidth]{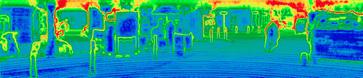}}
    \caption{Predicted images}
    \label{results_1:a}
    \end{subfigure}%
\caption{Uncertainty Analysis on Kitti depth completion selected validation set.
(a) Predicted dense-depth RMSE variance as a function keep ratio using our proposed method.
(b) Comparison of the predicted depth RMSE as a function of keep ratio between Nconv's confidence-map \cite{eldesokey2018confidence} denoted by blue bars, our proposed error-map denoted by red bars and our implementation of Aleatoric uncertainty-map \cite{kendall2017uncertainties} denoted by yellow bars.
(c) First image: Raw lidar + RGB image. Second image: Nconv' confidence-map. Third image: Aleatoric uncertainty-map. Fourth Image: Our error-map. Maps are represented in log space and values are shown using "nipy spectral" spectrum.
}
\vspace{-1.5em}
\label{uncertainak}
\end{center}
\end{figure*}
\section{Experimental Results}
\subsection{Kitti Depth Completion Dataset}
The Kitti depth completion dataset has 92,750 sparse depth (projected from Velodyne HDL-64E to the left and right cameras) along with semi-dense ground truth (generated by enforcing consistency between aggregated LiDARs and stereo depths). The dataset includes 1,000 selected validation images and 1’000 test images. Semi-dense ground-truth covers roughly 30\% of each image.
We trained our network using a batch size of 10 on two V100 GPUs. We used Adam optimizer with the learning rate of 0.001. We used the images in their original size for training (352x1216) and trained the model for 4 epochs.

To verify the proposed approach, we do the comparisons with the state-of-the-art approaches on Kitti depth completion benchmark leaderboard. Furthermore, we conduct uncertainty and error analysis of the predicted depth map. 
\subsubsection{Comparison with other methods}
Each ground-truth image in Kitti depth completion selected validation set on average contains 73,128 points. Figure \ref{var_all}(c) shows the predicted results of our network in 2D space. Visual analysis of the predictions in 2D space may be misleading since we don't see the real position of the points in 3D. We can observe apparent prediction errors on sparse areas, around the edges and further away points by reconstructing the 3D views from the predicted dense depth images as shown in Figure \ref{results_1}(c). Further, the reconstructed 3D point-clouds of other methods are illustrated in Figure~\ref{close_view}(c-d). Our analyses of the 3D point-clouds show the average error around the described points is approximately 100 times more than the average error on other parts. We remove points if they predicted error is higher than a predefined threshold $T$ (which represent the absolute error tolerance and is set by the user based on the application) and keep them otherwise. The percentage of the high-confidence predictions, which were kept after filtering out points with high errors is denoted by "Keep Ratio." 

A comparison with other methods appeared on the Kitti depth completion benchmark leaderboard is given in Table 1. As an example, our result outperforms the state-of-the-art algorithm without using RGB guidance by removing 4 points out of 73,128 (keep ratio $\approx$ 99.995$\%$), which contain significant errors. Besides, our result also outperforms the state-of-the-art algorithm using RGB guidance by removing 220 points out of 73,128 (keep ratio $\approx$ 99.7$\%$). Removing 731 points out of 73,128 (keep ratio $\approx$ 99\%) from our prediction yields to 145 mm improvement regarding RMSE from the best available algorithm (e.g., RGB guide and Certainty \cite{van2019sparse}) with RGB guidance.

Available methods applied on Kitti depth completion dataset exhibit a long tail error distribution and the error has a large variance. This justifies why removing a small portion of the predicted dense depths can yield to substantial improvement on RMSE of the predicted depths. Figure~\ref{uncertainak}(a) illustrates the variance of the predicted depth RMSE as a function of the keep ratio using our predicted error-map. 
We believe the effect of our method on the final error is more than what is described in Table 1. As we see in Figures~\ref{close_view}(b) and Figure~\ref{results_1}(b), most of the points lying on the sparse areas, edges and also far away points are removed from Kitti depth completion ground truth. However, all available methods are filling these areas and they do not get penalize since no ground truth is available for them.

\begin{table}[thb]
\begin{center}
\vspace{-0.5em}
\label{table:uncertain_table}
\caption{Comparison with other methods based on predicted error-map on Kitti selected validation set. Using our method users can choose their expected error threshold which yields to significantly lower prediction error.}
\scalebox{0.75}{
\begin{tabular}{ |c|c|c|c| }
\hline
\multicolumn{2}{|c|}{Methods}  &\begin{tabular}{@{}c@{}}Keep \\ Ratio (\%)\end{tabular} & \begin{tabular}{@{}c@{}}RMSE \\ (mm)\end{tabular} \\ \hline
\multirow{4}{*}{RGB guidance} & HMS-Net \cite{huang2018hms}& 100.00 & 883.74 \\
 & NConv-CNN-L2 \cite{eldesokey2018confidence}& 100.00  & 870.82\\
 & Sparse-to-Dense \cite{ma2018self}& 100.00     & 878.56 \\
 & RGB guide and Certainty \cite{van2019sparse} & 100.00  & \textbf{802.00} \\ \hline
\multirow{5}{*}{No RGB guidance} & Sparse ConvNet \cite{uhrig2017sparsity} & 100.00    & 1820.00 \\
 & ADNN \cite{chodosh2018deep}& 100.00    & 1350.00\\ 
 & IP-Basic \cite{ku2018defense} & 100.00  & 1350.93\\
 & Glob guide and Certainty\cite{van2019sparse} & 100.00  & 995.00 \\
 & HMS-Net \cite{huang2018hms}& 100.00 & \textbf{994.14}\\ \hline
\multirow{8}{*}{\begin{tabular}{@{}c@{}}Ours \\ (No RGB guidance)\end{tabular}} & threshold T: none & 100.00 &1004.28 \\
 & threshold T: 19140mm & 99.995  & \textbf{993.5} \\
 & threshold T: 4687mm& 99.7  & \textbf{800.00}  \\
 & threshold T: 2343mm& 99 & 657.77  \\
 & threshold T: 585mm& 95 & \textbf{399.03} \\
 & threshold T: 351mm & 90 & 300.00  \\
 & threshold T: 171mm& 80  & 208.01 \\
 & threshold T: 74mm & 50  & 129.94\\
\hline
\end{tabular}}
\vspace{-1.5em}
\end{center}
\end{table}

\begin{table}
\caption{Effects of adding foreground/background, our "Error Prediction" unit and Aleatoric uncertainty loss function to our base network with respect to predicted depth RMSE.}
\label{ablation_table}
\vspace{-0.5em}
\begin{center}
\begin{small}
\begin{sc}
\scalebox{0.61}{
\begin{tabular}{|c | c | c | c | c |}
\hline
  &\begin{tabular}{@{}c@{}}Our Base mdoel \\(no error prediction)\end{tabular}   &\begin{tabular}{@{}c@{}}Base + our\\"Error Prediction"\end{tabular}  &\begin{tabular}{@{}c@{}}Base + \\ Aleatoric\\ uncertainty\end{tabular} &\begin{tabular}{@{}c@{}}Base with\\ no fore/back- \\ -ground\end{tabular}     \\ \hline
\begin{tabular}{@{}c@{}} Depth\\ RMSE\end{tabular}      & 1004 & 1004 & 1130 & 1129 \\
\hline
\end{tabular} 
}
\end{sc}
\end{small}
\end{center}
\vspace{-2.5em}
\vskip -0.05in
\end{table}
\begin{figure*}
	\centering
	\begin{subfigure}[b]{0.166\textwidth}
		\includegraphics[width=2.5cm,height=1.7cm]{figures/nyu/00301_rgb.png}
		\includegraphics[width=2.5cm,height=1.7cm]{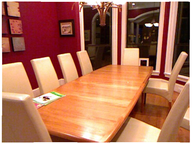}
		\caption{\begin{tabular}{@{}c@{}}RGB\\ image  \end{tabular}}
		\label{fig:gull}
	\end{subfigure}%
	\begin{subfigure}[b]{0.166\textwidth}
		\centering
		\includegraphics[width=2.5cm,height=1.7cm]{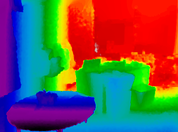}
		\includegraphics[width=2.5cm,height=1.7cm]{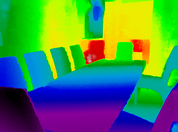}
		\caption{\begin{tabular}{@{}c@{}}Depth\\ ground-truth \end{tabular}}
		\label{fig:gull}
	\end{subfigure}%
	\begin{subfigure}[b]{0.166\textwidth}
		\centering
		\includegraphics[width=2.5cm,height=1.7cm]{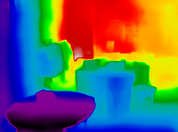}
		\includegraphics[width=2.5cm,height=1.7cm]{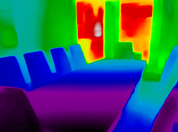}
		\caption{\begin{tabular}{@{}c@{}}Predicted\\ depth \end{tabular}}
		\label{fig:gull}
	\end{subfigure}%
	\begin{subfigure}[b]{0.166\textwidth}
		\centering
		\includegraphics[width=2.5cm,height=1.7cm]{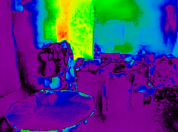}
		\includegraphics[width=2.5cm,height=1.7cm]{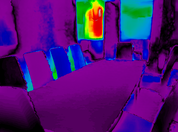}
		\caption{\begin{tabular}{@{}c@{}}Error-map\\ ground-truth \end{tabular}}
		\label{fig:gull}
	\end{subfigure}%
	\begin{subfigure}[b]{0.166\textwidth}
		\centering
		\includegraphics[width=2.5cm,height=1.7cm]{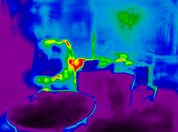}
		\includegraphics[width=2.5cm,height=1.7cm]{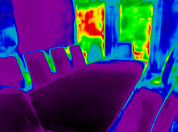}
		\caption{\begin{tabular}{@{}c@{}}Predicted\\ error-map \end{tabular}}
		\label{fig:gull}
	\end{subfigure}%
	\begin{subfigure}[b]{0.166\textwidth}
		\centering
		\includegraphics[width=2.5cm,height=1.7cm]{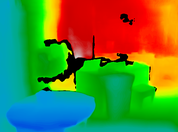}
		\includegraphics[width=2.5cm,height=1.7cm]{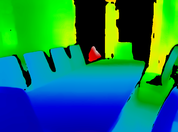}
		\caption{\begin{tabular}{@{}c@{}}After removing\\ high errors \end{tabular}}
		\label{fig:gull}
	\end{subfigure}%
	\caption{Error-map prediction for monocular depth estimation on NYU Depth dataset. (a) RGB image (b) Depth ground truth, (c) predicted depth-map, (d) error-map ground truth (absolute depth prediction error), (e) predicted error-map and (f) predicted semi-dense depth-map after removing points with predicted high error. Images colorization is based on each image relative pixel values.}
	\vspace{-1.0em}
	\label{NYU_all}
	
\end{figure*}
\subsubsection{Uncertainty Analysis}
In this section, we analyze the predicted error, confidence, and uncertainty maps with respect to the proposed approach, Nconv \cite{eldesokey2018confidence,eldesokey2018propagating}, and Aleatoric uncertainty in \cite{kendall2017uncertainties}. In order to find a unifying way to compare uncertainty, confidence and error maps, we did the comparison based on the "keep ratio" described in Section III. Since the quality of the predicted depth influences the predicted error-map (error-map is supposed to estimate the prediction error) it is not fare to compare different approaches based on MAE or RMSE of the error-maps and the ground-truth error. Our experimental results show that Nconv \cite{eldesokey2018confidence,eldesokey2018propagating} is not able to provide a precise confidence-map. Their confidences are generated using normalized convolutions and are correlated with the density of valid points. The confidence of LiDAR-based depth completion in 2D space have a little correlation with the density of valid points, since in 2D image space far away points look denser, while they are more sparse in 3D. LiDAR-based sparse depth also has a higher density in the warped areas described in Section III which increases the ambiguity. Both of these scenarios yield to lower confidence in the process of depth completion although \cite{eldesokey2018confidence,eldesokey2018propagating} predict high confidence for them. Using their approach increasing the confidence threshold to remove more than 20\% of the predicted dense depth, increases the RMSE error since we end up with far away points.

Aleatoric uncertainty described in \cite{kendall2017uncertainties} have not been used for depth completion prior to our work. We modified our loss function to predict the dense depth and the Aleatoric uncertainty. Following our "error prediction" unit structure, the first stream will predict the dense depth (mean) and the second stream predicts the uncertainty (noise/variance). The loss function is defined as below 
\vspace{-0.5em}
\begin{center}
\begin{equation}
Loss = \frac{\mid\mid y-\hat{y} \mid\mid^2}{2\sigma^2 } + \frac{log\sigma^2}{2}
\label{var_loss}
\end{equation}
\end{center}
\vspace{-0.5em}

with $\sigma$, $y$, and $\hat y$ denoting the uncertainty, dense depth, and depth ground truth respectively. Authors in \cite{kendall2017uncertainties} predict the \textit{log} of the error, add softplus layer for uncertainty, and optimize their model for MAE not MSE for stability. Still, the training process is quite unstable and slow. To get comparable results with our method, we had to tweak our network architecture (since it did not converge well using our foreground/background approach) and continue training the model 5 times more than our approach. We think this is because network needs to find a way to tune two predictions of $log\sigma$ and $y$ without any guidance. Our method does not face this issue because of the guided training and completely separate objective functions for the predicted depth and error-map. In brief, our approach is different from \cite{kendall2017uncertainties} in following aspects: 
\begin{enumerate}
\item Authors in \cite{kendall2017uncertainties} propose a loss function which forces the uncertainty and depth to have a same type. As an example in NYU Depth dataset we learned the depth using inverse depth values and at the same time learned the error-map using original depth values. It is not possible to have the same approach using \cite{kendall2017uncertainties}'s loss function. We further discuss this in the next Section.
\item Authors in \cite{kendall2017uncertainties} predict the uncertainty by probabilistic interpretation of deep networks. We directly estimate the error by learning the error signal.
\item Our predicted error has a valid contextual meaning which is the estimate error in that specific pixel. Authors in \cite{kendall2017uncertainties} predict the uncertainty not the error-map.
\item Authors in \cite{kendall2017uncertainties} propose a loss function which need both logarithm and division of the predicted outputs which makes the training unstable, slow and hard to converge.  
\end{enumerate}
The ablation study with respect to RMSE of the predicted depth is presented in Table~\ref{ablation_table}.
 Figure~\ref{uncertainak}(c) shows an example of the predicted confidence, uncertainty and error-map using Nconv, \cite{kendall2017uncertainties}'s Aleatoric uncertainty and our proposed method. Figure~\ref{uncertainak}(b) shows the comparison between the described methods regarding the root mean squared error as a function of keep ratio percentage.
 Our results based on Aleatoric uncertainty and our proposed error-map significantly outperform Nconv. As described in Section IV, error of depth completion methods on Kitti has a very high variance although Equation 6 ignores points with high errors by decreasing their training weight and focuses on points that it can predict accurately. Therefore, using \cite{kendall2017uncertainties}'s Aleatoric uncertainty we are not able to get a comparable accuracy regarding the RMSE and MAE with our approach without removing any points, although results get slightly better than our method as we decrease the keep ratio percentage less than 95\%. This is because of the fact that predicted depth values using Aleatoric uncertainty are biased toward the points with lower error.

\begin{table}[thb]
\begin{center}
\vspace{-0.5em}
\label{table:uncertain_table_nyu}
\caption{Comparison with other methods based on predicted error-map on NYU depth Eigen \cite{eigen2014depth} split test set. Using our method users can choose their expected error threshold which yields to significantly lower prediction error.}
\scalebox{0.68}{
\begin{tabular}{ |c|c|c|c|c|c|c|c|c| }
\hline
\multicolumn{2}{|c|}{Methods}  &\begin{tabular}{@{}c@{}}Keep \\ Ratio (\%)\end{tabular} & $\delta_1\uparrow$ & $\delta_2\uparrow$ & $\delta_3\uparrow$ & $rel\downarrow$ &$\textbf{rms}\downarrow$ &$log_{10}\downarrow$ \\ \hline
\multirow{7}{*}{\begin{tabular}{@{}c@{}}Other\\methods\end{tabular}}
 & Eigen et al. \cite{eigen2014depth} & 100.00 &0.769  & 0.950 & 0.988 & 0.158 & 0.641 & -\\
 & Laina et al. \cite{laina2016deeper}& 100.00 &0.811 & 0.953&0.988 & 0.127 & 0.573 & 0.055 \\
 & MS-CRF \cite{xu2017multi}& 100.00 &0.811 & 0.954 & 0.987 & 0.121 & 0.586 & 0.052 \\ 
 & Hao et al. \cite{hao2018detail}& 100.00  & 0.841 & 0.966 & 0.991 & 0.127 & 0.555 & 0.053 \\ 
 & \begin{tabular}{@{}c@{}}Alhashim et al. \cite{alhashim2018high} \\ (Github model)\end{tabular} & 100.00  &\textbf{0.847} & \textbf{0.973} & \textbf{0.994} & 0.127 & 0.548 & 0.053 \\ 
 & Fu et al. \cite{fu2018deep}& 100.00  &0.828 & 0.965 & 0.992 & \textbf{0.115} & \textbf{0.509} & \textbf{0.051} \\
\hline

\multirow{7}{*}{\begin{tabular}{@{}c@{}} Alhashim\\et al. \cite{alhashim2018high}\\+\\Our\\Error Unit\end{tabular}} 
 & threshold T: none & 100.00 & 0.843 & 0.971 & 0.993 & 0.129 & 0.549 & 0.053  \\
 & threshold T: 350mm& 98.44 & \textbf{0.847} & \textbf{0.973} & \textbf{0.994} & 0.126 & 0.516 & 0.053  \\
 & threshold T: \textbf{300mm}& \textbf{97.53} & 0.849 & 0.974 & 0.994 & 0.126 & \textbf{0.504} & 0.053  \\
 & threshold T: 250mm & 96.05 & 0.852 & 0.975 & 0.995 & 0.125 & 0.489 & 0.053 \\
 & threshold T: 200mm& 93.52 & 0.856 & 0.976 & 0.995 & 0.123 & 0.468 & 0.052 \\
 & threshold T: 100mm& 76.37 & 0.871 & 0.9817 & 0.997 & 0.116 & 0.365 & \textbf{0.049} \\
 & threshold T: 70mm& 59.39 & 0.874 & 0.984 & 0.998 & \textbf{0.114} & 0.319 & 0.048 \\
\hline
\end{tabular}}
\vspace{-1.5em}
\end{center}
\end{table}
\subsection{Monocular Depth Estimation on NYU Depth dataset}
In order to demonstrate the ability of the proposed method to be extended on other regression tasks we did further experiments on monocular depth estimation on NYU Depth dataset. We followed the method proposed by \cite{alhashim2018high} and added our error prediction unit to their algorithm without any further tuning of the algorithm. Authors in \cite{alhashim2018high} train their convolutional neural network by adding 3 different losses: MAE for inverse depth values, gradient loss of the depth image and Structural Similarity SSIM loss \cite{wang2004image}. They have used hand engineered weights for each of the losses. We simply replaces the hand engineered weights with our normalization technique (ratio loss) described in section III and added our error-map loss to their algorithm. We also tried to add Aleatoric uncertainty to their algorithm but since they optimize for inverse depth we only can estimate the inverse depth's uncertainty which will not represent the true uncertainty of the trained model (since it will be lower for far away points). The results are represented in Figure~\ref{NYU_all}. The comparison with other state-of-the-art approaches in Monocular Depth Estimation are represented in Table III.
\vspace{-0.5em}
\section{Conclusion}
In this work, we introduce a new method to train a deep convolutional network to simultaneously predict depth and error-map with no degradation regarding the depth prediction, in a multi-tasking manner. Using the proposed method we investigate the process of generating dense point-cloud using the sparse point-cloud of LiDAR sensors. Our method is straightforward to implement and outperforms the available methods for depth uncertainty and confidence.
{
\small
\bibliographystyle{IEEEtran}
\bibliography{egbib}
}



\section*{APPENDIX}

\subsection{Our VLP-32C Data}
\begin{figure*}
	\centering
	\begin{subfigure}[b]{0.33\textwidth}
		\centering
		\includegraphics[width=\columnwidth,height=0.45\columnwidth]{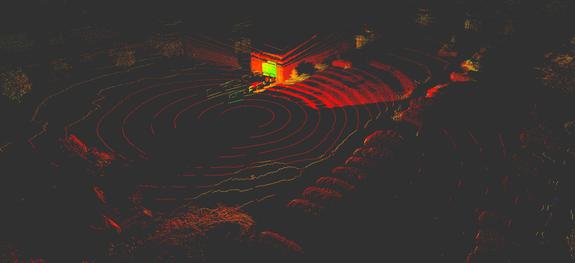}
		\caption{\begin{tabular}{@{}c@{}}Depth completion applied\\only on front view  \end{tabular} }
		\label{fig:gull}
	\end{subfigure}%
	\begin{subfigure}[b]{0.33\textwidth}
		\centering
		\includegraphics[width=\columnwidth,height=0.45\columnwidth]{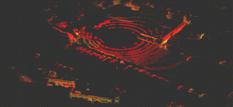}
		\caption{\begin{tabular}{@{}c@{}}Error threshold\\ 585  \end{tabular}  }
		\label{fig:gull}
	\end{subfigure}%
	\begin{subfigure}[b]{0.33\textwidth}
		\centering
		\includegraphics[width=\columnwidth,height=0.45\columnwidth]{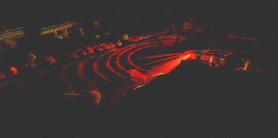}
		\caption{\begin{tabular}{@{}c@{}}Error threshold\\ 390  \end{tabular} }
		\label{fig:gull}
	\end{subfigure}%
	\caption{Our approach applied to VLP-32C. First image: Depth completion only applied on front view (Error threshold: 585mm). Second image: 360 depth completion (Error threshold: 585mm). Third image: 360 depth completion (Error threshold: 390mm).}
	\label{dense_compare}
	
\subsection{Complementary Results}
	
\end{figure*}

We also used our VLP-32C lidar data to reconstruct the 3D view. The spare 3D point-cloud is projected onto 5 virtual cameras. Since we trained our network using Kitti each camera will have 80$^\circ$ filed of view (FOV) looking around the vehicle to cover 360$^\circ$ of FOV. We pass five 2D sparse depth images through our CNN to get the dense depth images and their corresponding error-map. Hence the error of depth completion increases by moving from 64 to 32 beams of LiDAR, filtering the predicted depth based on the predicted error-map is essential. In the last step, we project the 2D images back to the 3D view. The result is shown in Figure~\ref{dense_compare}.   
\begin{figure*}
	\centering
	\begin{subfigure}[b]{0.2\textwidth}
		\centering
		\includegraphics[width=4.2cm,height=3cm]{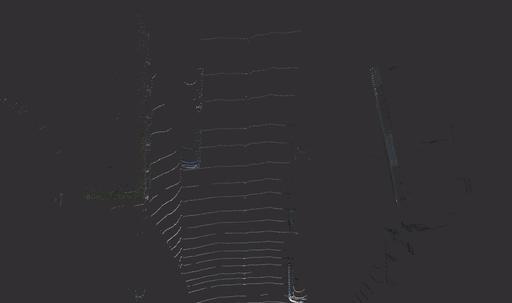}        
		\includegraphics[width=4.2cm,height=3cm]{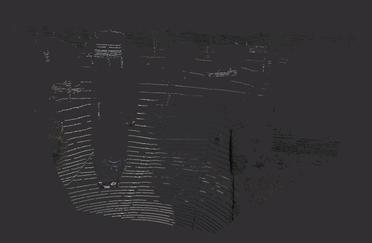}
		\includegraphics[width=4.2cm,height=3cm]{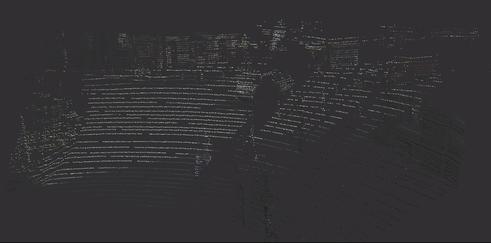}
		
		\caption{\begin{tabular}{@{}c@{}}Raw LiDAR\\ Kitti\end{tabular}}
		\label{fig:gull}
	\end{subfigure}%
	\begin{subfigure}[b]{0.2\textwidth}
		\centering
		\includegraphics[width=4.2cm,height=3cm]{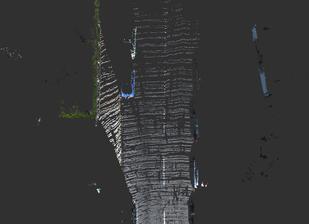}
		\includegraphics[width=4.2cm,height=3cm]{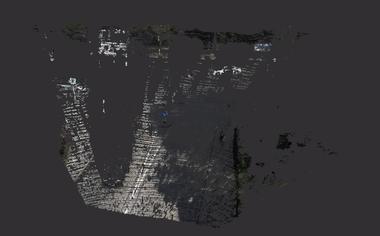}
		\includegraphics[width=4.2cm,height=3cm]{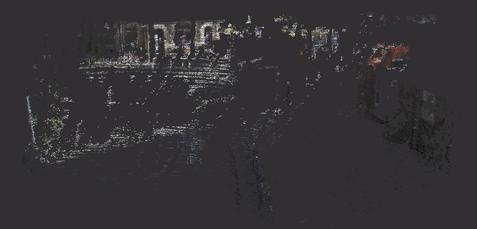}
		\caption{\begin{tabular}{@{}c@{}}Ground truth\\ Kitti\end{tabular}}
		\label{fig:gull}
	\end{subfigure}%
	\begin{subfigure}[b]{0.2\textwidth}
		\centering
		\includegraphics[width=4.2cm,height=3cm]{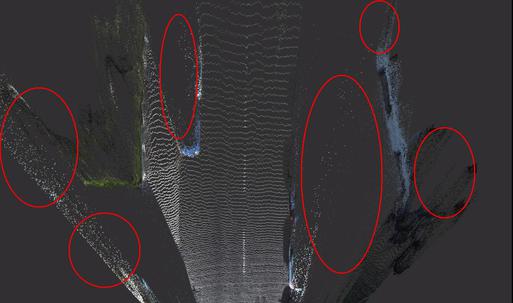}
		\includegraphics[width=4.2cm,height=3cm]{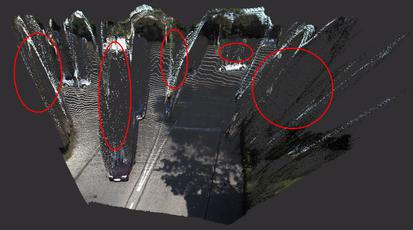}
		\includegraphics[width=4.2cm,height=3cm]{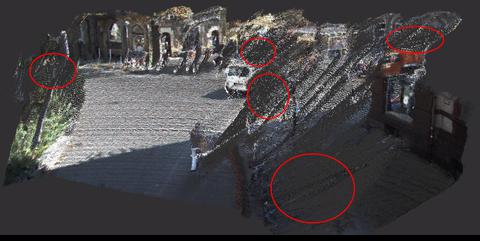}
		\caption{\begin{tabular}{@{}c@{}}Prediction without\\ removing errors \end{tabular} }
		\label{fig:gull}
	\end{subfigure}%
	\begin{subfigure}[b]{0.2\textwidth}
		\centering
		\includegraphics[width=4.2cm,height=3cm]{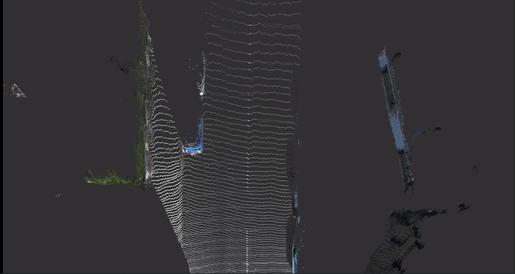}
		\includegraphics[width=4.2cm,height=3cm]{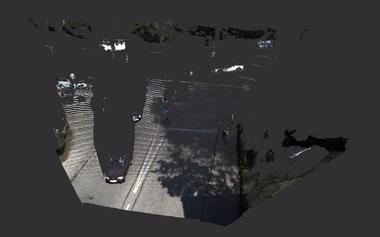}
		\includegraphics[width=4.2cm,height=3cm]{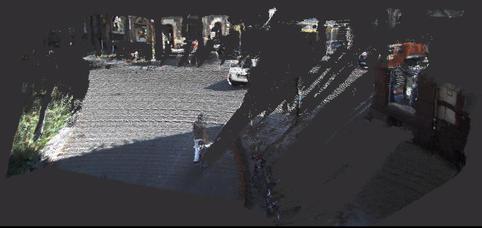}
		\caption{\begin{tabular}{@{}c@{}}After removing\\ predicted high errors \end{tabular}}
		\label{fig:gull}
	\end{subfigure}%
	\caption{Comparison between 3D point-clouds colored with RGB values.(a) Raw LiDAR point-cloud, (b) Kitti ground-truth, (c) Predicted dense point-cloud, (d) Predicted point-cloud after removing points with predicted high errors.}  
	\label{ALLLL_2}
\end{figure*}
\begin{figure*}

	\centering
	\begin{subfigure}[b]{0.166\textwidth}
		\centering
		\includegraphics[width=3cm,height=2.2cm]{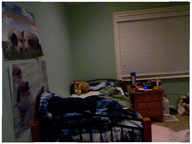}
		\includegraphics[width=3cm,height=2.2cm]{figures/nyu/00238_rgb.png}
		\includegraphics[width=3cm,height=2.2cm]{figures/nyu/00301_rgb.png}
		\includegraphics[width=3cm,height=2.2cm]{figures/nyu/00329_rgb.png}
		\includegraphics[width=3cm,height=2.2cm]{figures/nyu/00357_rgb.png}
		\includegraphics[width=3cm,height=2.2cm]{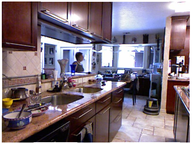}
		\includegraphics[width=3cm,height=2.2cm]{figures/nyu/00525_rgb.png}
		\includegraphics[width=3cm,height=2.2cm]{figures/nyu/00539_rgb.png}
		\includegraphics[width=3cm,height=2.2cm]{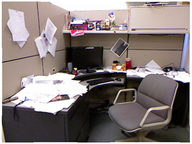}
		\includegraphics[width=3cm,height=2.2cm]{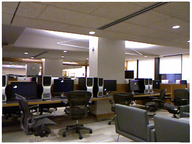}
		\caption{\begin{tabular}{@{}c@{}}RGB\\ image  \end{tabular}}
		\label{fig:gull}
	\end{subfigure}%
	\begin{subfigure}[b]{0.166\textwidth}
		\centering
		\includegraphics[width=3cm,height=2.2cm]{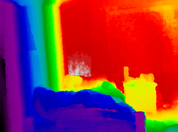}
		\includegraphics[width=3cm,height=2.2cm]{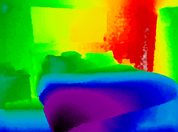}
		\includegraphics[width=3cm,height=2.2cm]{figures/nyu/00301_gt.png}
		\includegraphics[width=3cm,height=2.2cm]{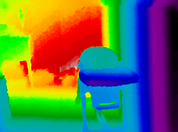}
		\includegraphics[width=3cm,height=2.2cm]{figures/nyu/00357_gt.png}
		\includegraphics[width=3cm,height=2.2cm]{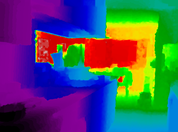}
		\includegraphics[width=3cm,height=2.2cm]{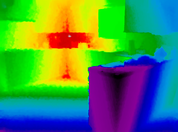}
		\includegraphics[width=3cm,height=2.2cm]{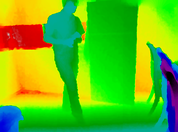}
		\includegraphics[width=3cm,height=2.2cm]{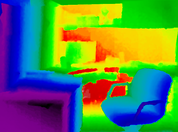}
		\includegraphics[width=3cm,height=2.2cm]{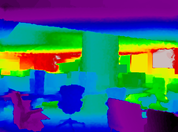}
		\caption{\begin{tabular}{@{}c@{}}Depth\\ ground-truth \end{tabular}}
		\label{fig:gull}
	\end{subfigure}%
	\begin{subfigure}[b]{0.166\textwidth}
		\centering
		\includegraphics[width=3cm,height=2.2cm]{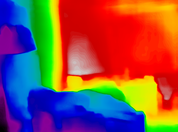}
		\includegraphics[width=3cm,height=2.2cm]{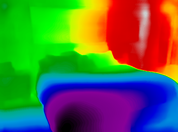}
		\includegraphics[width=3cm,height=2.2cm]{figures/nyu/00301_depth.png}
		\includegraphics[width=3cm,height=2.2cm]{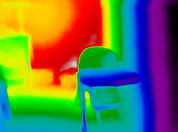}
		\includegraphics[width=3cm,height=2.2cm]{figures/nyu/00357_depth.png}
		\includegraphics[width=3cm,height=2.2cm]{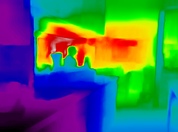}
		\includegraphics[width=3cm,height=2.2cm]{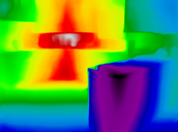}
		\includegraphics[width=3cm,height=2.2cm]{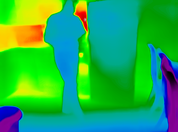}
		\includegraphics[width=3cm,height=2.2cm]{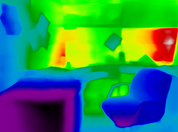}
		\includegraphics[width=3cm,height=2.2cm]{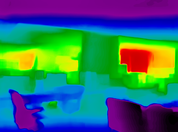}
		\caption{\begin{tabular}{@{}c@{}}Predicted\\ depth \end{tabular}}
		\label{fig:gull}
	\end{subfigure}%
	\begin{subfigure}[b]{0.166\textwidth}
		\centering
		\includegraphics[width=3cm,height=2.2cm]{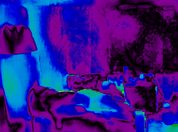}
		\includegraphics[width=3cm,height=2.2cm]{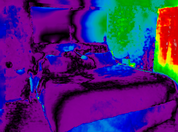}
		\includegraphics[width=3cm,height=2.2cm]{figures/nyu/00301_error_gt.png}
		\includegraphics[width=3cm,height=2.2cm]{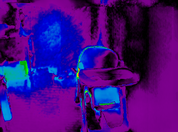}
		\includegraphics[width=3cm,height=2.2cm]{figures/nyu/00357_error_gt.png}
		\includegraphics[width=3cm,height=2.2cm]{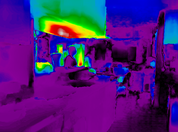}
		\includegraphics[width=3cm,height=2.2cm]{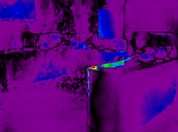}
		\includegraphics[width=3cm,height=2.2cm]{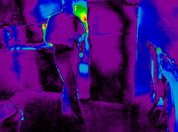}
		\includegraphics[width=3cm,height=2.2cm]{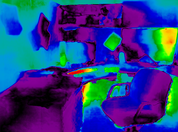}
		\includegraphics[width=3cm,height=2.2cm]{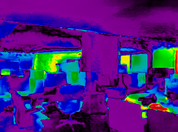}
		\caption{\begin{tabular}{@{}c@{}}Error-map\\ ground-truth \end{tabular}}
		\label{fig:gull}
	\end{subfigure}%
	\begin{subfigure}[b]{0.166\textwidth}
		\centering
		\includegraphics[width=3cm,height=2.2cm]{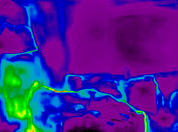}
		\includegraphics[width=3cm,height=2.2cm]{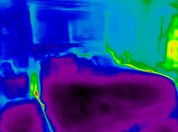}
		\includegraphics[width=3cm,height=2.2cm]{figures/nyu/00301_error.png}
		\includegraphics[width=3cm,height=2.2cm]{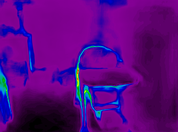}
		\includegraphics[width=3cm,height=2.2cm]{figures/nyu/00357_error.png}
		\includegraphics[width=3cm,height=2.2cm]{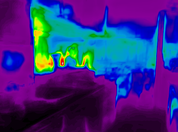}
		\includegraphics[width=3cm,height=2.2cm]{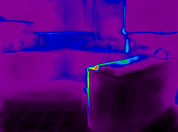}
		\includegraphics[width=3cm,height=2.2cm]{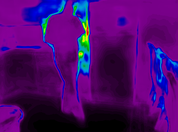}
		\includegraphics[width=3cm,height=2.2cm]{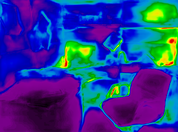}
		\includegraphics[width=3cm,height=2.2cm]{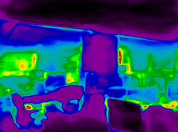}
		\caption{\begin{tabular}{@{}c@{}}Predicted\\ error-map \end{tabular}}
		\label{fig:gull}
	\end{subfigure}%
	\begin{subfigure}[b]{0.166\textwidth}
		\centering
		\includegraphics[width=3cm,height=2.2cm]{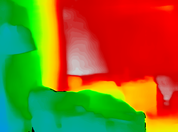}
		\includegraphics[width=3cm,height=2.2cm]{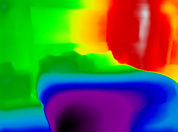}
		\includegraphics[width=3cm,height=2.2cm]{figures/nyu/00301_removed.png}
		\includegraphics[width=3cm,height=2.2cm]{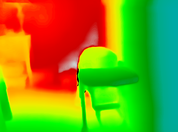}
		\includegraphics[width=3cm,height=2.2cm]{figures/nyu/00357_removed.png}
		\includegraphics[width=3cm,height=2.2cm]{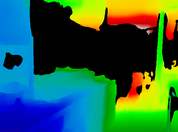}
		\includegraphics[width=3cm,height=2.2cm]{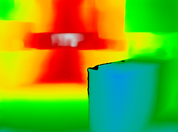}
		\includegraphics[width=3cm,height=2.2cm]{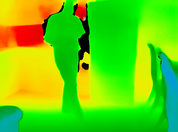}
		\includegraphics[width=3cm,height=2.2cm]{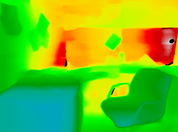}
		\includegraphics[width=3cm,height=2.2cm]{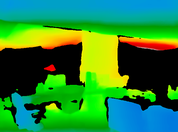}
		\caption{\begin{tabular}{@{}c@{}}After removing\\ high errors \end{tabular}}
		\label{fig:gull}
	\end{subfigure}%
	\caption{Error-map prediction for monocular depth estimation on NYU Depth dataset. (a) RGB image (b) Depth ground truth, (c) predicted depth-map, (d) error-map ground truth (absolute depth prediction error), (e) predicted error-map and (f) predicted semi-dense depth-map after removing points with predicted high error. Images colorization is based on each image relative pixel values.}
	\label{NYU_all_2}
\end{figure*}



\end{document}